\documentclass[letterpaper, 10 pt, conference]{ieeeconf}
\usepackage{cite}
\IEEEoverridecommandlockouts                              
\usepackage{amsmath}
\usepackage{graphicx}
\usepackage[caption=false]{subfig}
\usepackage{multirow}
\usepackage{dblfloatfix}
\usepackage{siunitx}
\usepackage{booktabs}
\usepackage{longtable}

\title{\LARGE \bf
DroneTrap: Drone Catching in Midair by Soft Robotic Hand with Color-Based Force Detection and Hand Gesture Recognition
}

\author{ 
Aleksey Fedoseev$^{1}$,
Valerii Serpiva$^{1}$,
Ekaterina Karmanova$^{1}$,
Miguel Altamirano Cabrera$^{1}$,
Vladimir Shirokun$^{1}$,\\
Iakov Vasilev$^{1}$,
Stanislav Savushkin$^{1}$,
Dzmitry Tsetserukou$^{1}$

\thanks{$^{*}$This work is funded by RFBR and CNRS, project number 21-58-15006.}
\thanks{$^{1}$The authors are with the Space Center, Skolkovo Institute of Science and Technology (Skoltech), 121205 Bolshoy Boulevard 30, bld. 1, Moscow, Russia. {\tt\small \{aleksey.fedoseev, valerii.serpiva, ekaterina.karmanova, miguel.altamirano, vladimir.shirokun, iakov.vasilev, stanislav.savushkin, d.tsetserukou\}@skoltech.ru}}%

}

\begin{document}

\maketitle
\thispagestyle{empty}
\pagestyle{empty}

\begin{abstract}

The paper proposes a novel concept of docking drones to make this process as safe and fast as possible. The idea behind the project is that a robot with a soft gripper grasps the drone in midair. The human operator navigates the robotic arm with the ML-based gesture recognition interface. 
The 3-finger robot hand with soft fingers is equipped with touch sensors, making it possible to achieve safe drone catching and avoid inadvertent damage to the drone's propellers and motors.  
Additionally, the soft hand is featured with a unique color-based force estimation technology based on a computer vision (CV) system. Moreover, the visual color-changing system makes it easier for the human operator to interpret the applied forces. 

Without any additional programming, the operator has full real-time control of robot's motion and task execution by wearing a mocap glove with gesture recognition, which was developed and applied for the high-level control of DroneTrap.

The experimental results revealed that the developed color-based force estimation can be applied for rigid object capturing with high precision (95.3\%). The proposed technology can potentially revolutionize the landing and deployment of drones for parcel delivery on uneven ground, structure maintenance and inspection, risque operations, and etc. 



\end{abstract}

\section{Introduction}



\begin{figure}[!h]
   \includegraphics[width=0.95\linewidth]{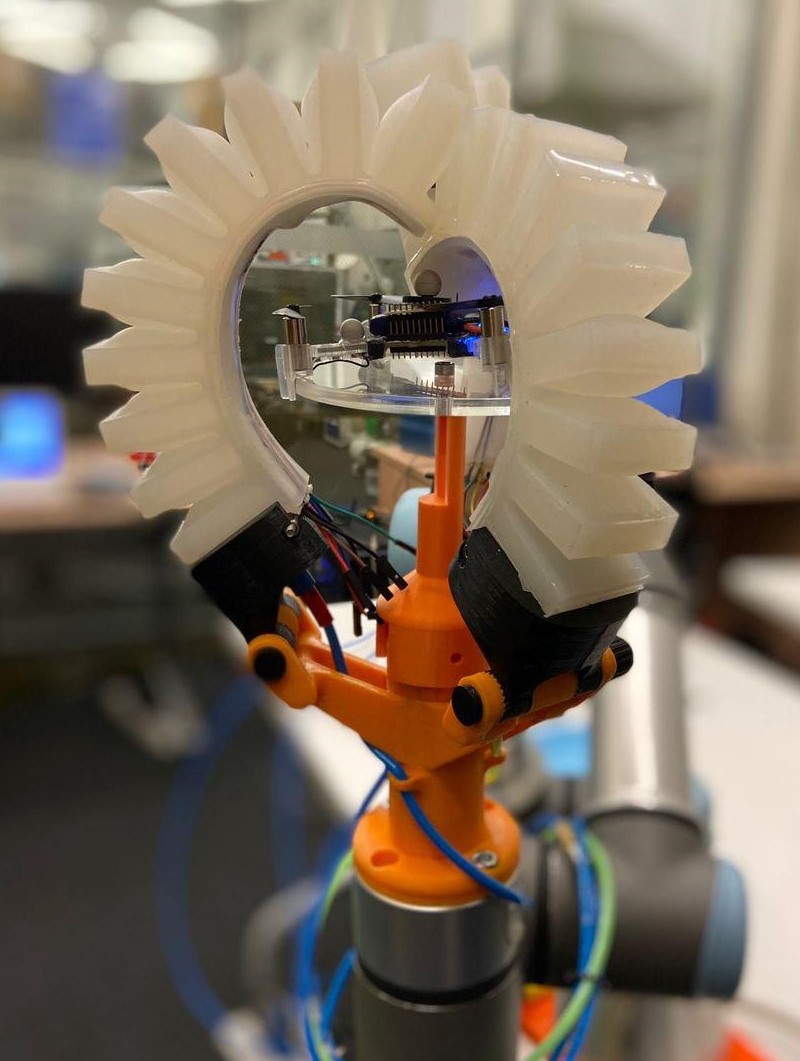}
  \caption{DroneTrap system performs a drone catching task by soft gripper.}
  \label{fig:main}
\end{figure}

Unmanned Aerial Vehicle (UAV) accidents mostly happen during landing, specifically in high winds or uneven terrain. The concept of landing the swarm on drones on the human arms for fast and safe deployment is proposed in \cite{Tsykunov}. However, it is expected that drones will be able to land in autonomous mode safely. Sarkisov et al. invented the landing gear with robotic legs to allow drones to adapt to the surface \cite{Sarkisov}. The legs are equipped with optical torque sensors making them responsive to the contact force with the ground.


Another approach to drone landing was presented by Miron et al. \cite{Miron}, proposing the drone-catching gripper based on sleeved bending actuators with a silicon membrane.
The drone landing system with 3 soft landing gears attached to the drone and neural network developed to achieve the desired landing behavior was introduced by Luo et al. \cite{Luo}. Such an approach allows high mobility for the UAV and operations on non-flat surfaces yet has a limited field of implementation since it decreases the weight of the possible drone payload. An origami-inspired cargo drone with a foldable cage is proposed by Kornatowski et al. \cite{Kornatowski_2017} for safe landing and interaction with humans.
Capturing a flying drone in midair with a net carried by cooperative UAVs is proposed by Rothe et al. \cite{Rothe}. A high-speed UAV parcel handover system with a high-speed vision system for the supply station control has been introduced by Tanaka et al. \cite{Tanaka_delivery}. This research also suggests a non-stop UAV delivery concept based on a high-speed visual control with a 6-axis robot.

The autonomous robotic systems were suggested in several researches on robotic hands \cite{Kim_Catching} and mobile robots \cite{Kao_omni}, for capturing the thrown objects with precise tracking. Several systems are focused on the precision of catching, which is demonstrated by the robots for juggling proposed by Kober et al. \cite{Kober_Juggling}. 
However, such systems do not take into account the fragility of the operated objects. Several researches propose either the position-based impedance controllers suggested for soft robotic catching of a falling object by Uchiyama et al. \cite{Uchiyama} or joint impedance tracking controller based on torque sensors, proposed by Bäuml et al. \cite{Bauml_flying_ball}. 

Another challenge for fragile flying vehicles is the spacecraft docking with the International Space Station (ISS). Canadarm2 with 7-DoF (Degrees of Freedom) captures and docks the unmanned cargo ships \cite{canadaarm}. It has a unique gripper with a three-wire crossover design featuring gentle grasping of such payload as satellites. The Canadarm2 key advantage is that it significantly reduces docking time and makes the process safer by avoiding crashes with ISS.

\section{Related Works}

\subsection{Soft gripper design}
There are several proposed gripper designs \cite{Murray_2015, Liang_2019, Sinatr},  achieving sufficient control over grasping.
However, in our work, we hypothesized that the three-fingered gripper based on PneuNets actuators proposed in \cite{Manns_2018} is sufficient to safely lock 3 contact points of a solid object in 3D space, and, therefore, can hold the drone securely.
Several approaches of embedding sensors into silicon and 3D-printed robotic grippers were introduced \cite{Thurutheleaav1488}, \cite{Shih}, allowing sufficient and safe robotic manipulations with fragile objects.
In this work, a force-sensitive resistor and linear flex sensor were embedded in each DroneTrap finger to achieve reliable and safe interaction.

\subsection{Gesture-based robot control}
There has been an increasing interest in applying Human-Robot Interaction (HRI) with gesture interface for controlling robots in addition to fully autonomous systems. This approach allows users to have adjustable control of robot motion with minimum delay and modify the task without additional programming, which plays a significant role in the midair dynamic teleoperation. One of the possible approach for such operation implements wearable body motion capturing (mocap) systems, e.g., exoskeletons \cite{8206145}, inertial measurement unit-based suits \cite{8764016}, and camera-based systems \cite{8172443}.
A glove controller was proposed by Fang et al. \cite{Fang_2018}, utilizing IMMU sensors to achieve simple and robust gesture recognition. To decrease the number of sensors and, hence, the complicity of wearable system, we developed a flexure sensor-based mocap glove as the gesture input interface.

We propose the technology DroneTrap to achieve a safe process of drone landing and deployment. Our novel system utilizes a gesture-based interface and color-based force detection to achieve a safe semi-autonomous manipulation with drones in midair for docking on uneven surfaces without complex control of the swarm landing. 


\section{ System Overview}
    
\begin{figure}[!h]
  \includegraphics[width=1.0\linewidth]{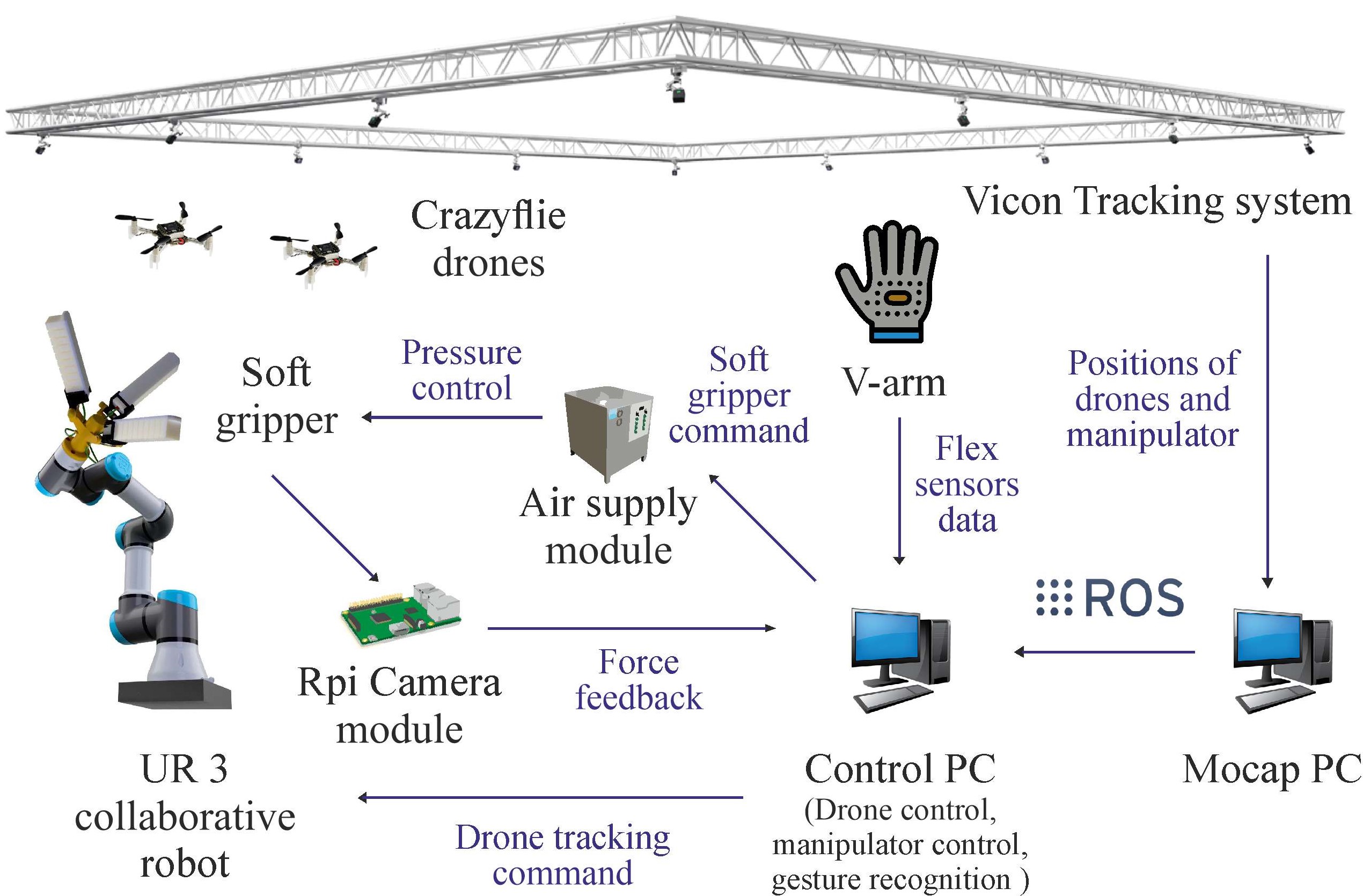}
  \caption{DroneTrap system architecture. Both the robotic arm and soft gripper are controlled with gesture recognition, allowing the operator to choose the desired sequence of drone catching.}
  \label{fig:SystemOverview}
\end{figure}

The drone  grasping experiments are conducted in the Laboratory facilitated with 12 cameras of Vicon V5 motion capturing system. This system can provide position and orientation information of the marked object (Crazyflie nano-quadrotors and reference frame of UR3 collaborative robot). Control PC with Unity framework receives positions from the Mocap PC by Robot Operating System (ROS) Kinetic.
Both the manipulator and the gripper are controlled by gesture recognition through the C\# framework, developed in Unity Engine. The custom mocap device V-Arm was designed to recognize the user's gestures from the flex sensor data. The proposed DroneTrap system is shown in Fig. \ref{fig:SystemOverview}.

\subsection{Soft gripper control}

\begin{figure}[htbp]
  \includegraphics[width=0.9\linewidth]{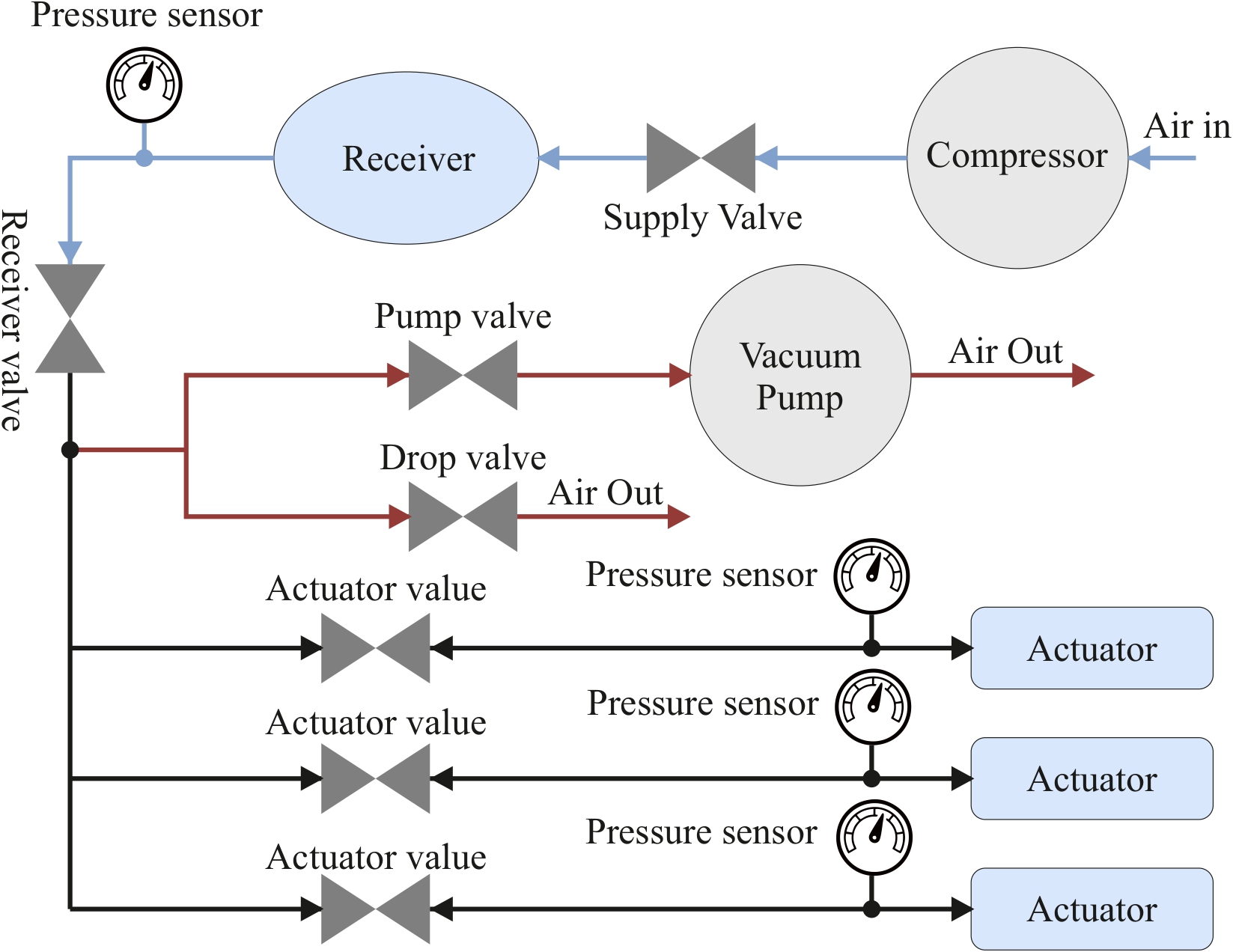}
  \caption{Control scheme for pneumatic actuators.}
  \label{fig:architecture}
\end{figure}

The flexible gripper consists of three SPAs (Soft Pneumatic Actuators) made of silicone rubber (Dragon Skin 30). SPAs are fixed on a special holder at 120 deg. to each other. Each of the actuators has a force sensor and a flexible sensor, the grip angle can be adjusted for different types of objects. A separate pneumatic hose is connected to each of the actuators. The compression and release of the gripper and each of the three pneumatic actuators are controlled by an Air  Supply  Module  (ASM)  with a pneumatic cylinder, pumps, and pneumatic valves (Fig. \ref{fig:architecture}).

Such architecture enables to inflate and deflate each SPA independently using minimum quantity of solenoid valves and pressure sensors. In this scheme, arrows indicate which directions air can flow (blue, red, and black colors represent supply,  exhaust, and bidirectional air flow, respectively).

\subsection{Gripper force estimation based on CV}

\begin{figure}[!h]
  \includegraphics[width=0.9\linewidth]{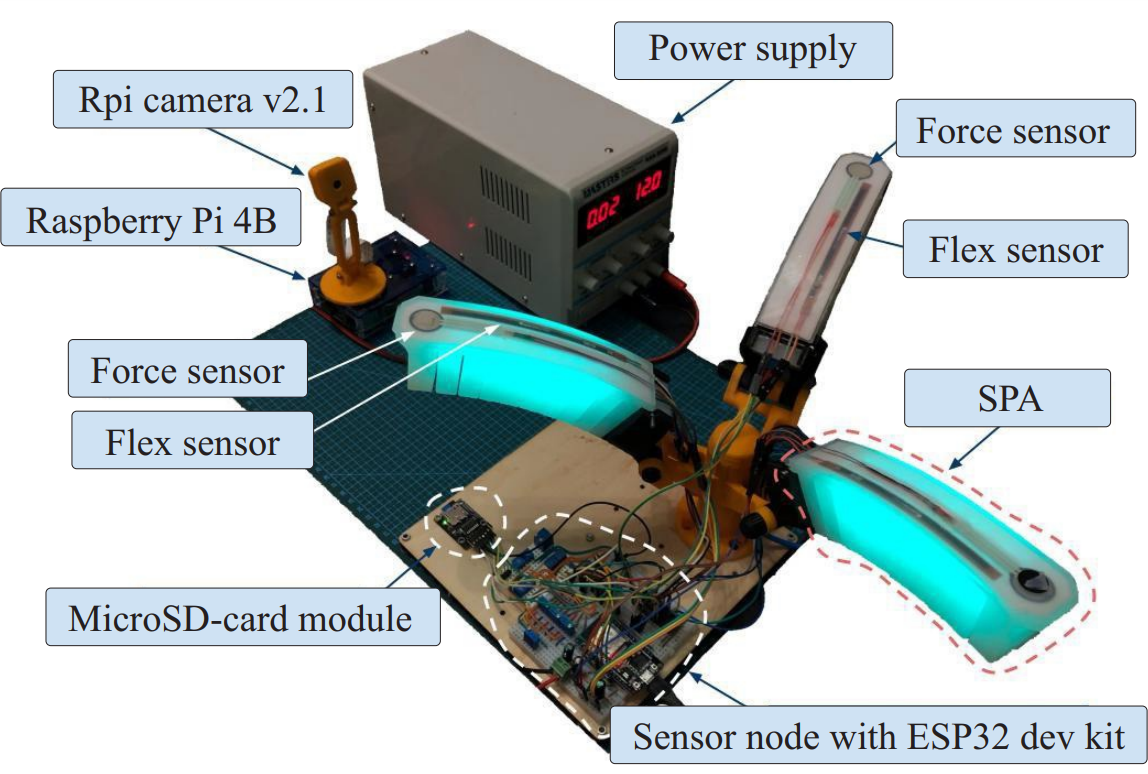}
  \caption{Soft gripper design and Computer Vision (CV) module.}
  \label{fig:setup}
\end{figure}

The color-based CV module was developed for gripper force evaluation, allowing to naturally recognize and control the applying force both autonomously and in supervised teleoperation. This module is responsible for gripper color recognition and data representation. It consists of two main components: a single-board computer Raspberry Pi 4B for data processing and  RPi Camera module v2.1. SPA gripper, sensor node module, camera, and data processing module are presented in Fig. \ref{fig:setup}. The camera is mounted 30 cm from the gripper, focused, and put at the same altitude.
    
\begin{figure}[!h]
    \centering
  \includegraphics[width=0.9\linewidth]{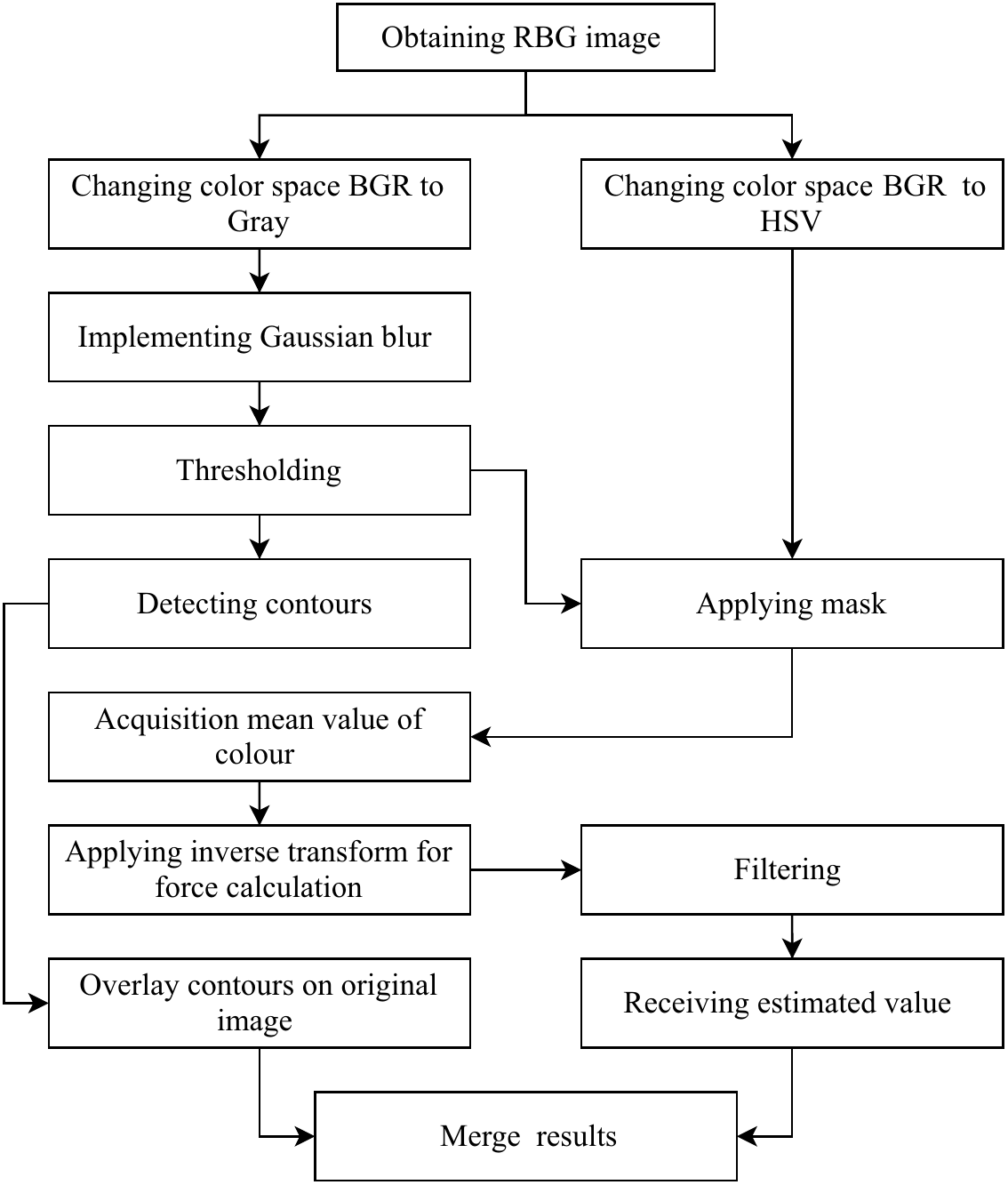}
  \caption{Force estimation algorithm with OpenCV.}
  \label{fig:color_b_CV}
\end{figure}

To estimate the force applied to human fingertips, the pixel brightness and HSV approaches have been proposed by Grieve et al. \cite{Grieve_2016} and Sartison et al. \cite{Sartison_2018}. In our research, we apply a HSV color model for a CV-based gripper force estimation. While in basic color models, hue level is represented either by 0 to 360-degree or  0 to 100 percent, the FastLED library was applied to decrease the data processing delay. The library was adjusted to HSV space instead of RGB, applying one-byte values (from 0 to 255) for hue, saturation, and value independently. To improve the estimation accuracy and avoid color duplication in the FastLED hue spectrum, the hue range was limited to 45-210 range, which corresponds to orange - purple colors. 
A core algorithm of the remote color-based applied force detection had been developed in Python 3.7 using the OpenCV library (Fig. \ref{fig:color_b_CV}).

Thresholding was used to convert an image to a binary array with implemented Otsu's method \cite{Otsu_1979}. 
For the contour detection a mask obtained in the previous step was applied to the original image. Then, contours were filtered by area size ($\textgreater$ 5000 px). A hue value conveyed to LED array is constructed of 2 components: maximum value of force sensitive resistor (FSR) readings (max value from 3 fingers of the gripper) and average value of 3 linear flex sensor (FS-L) readings (from all fingers). The sensor data, being stored in 12-bit register, was constrained in the range [45, 210] with a mapping Eq. (\ref{eq: mapping_reg}). 
Finally, the average value from sensors for the hue was determined in Eq. (\ref{eq: hue}):



\begin{equation}
\begin{cases}
 & R_{fsr}\: [0,\: 4096] \rightarrow Hue_{fsr}\: [45, \: 210]\\ 
 & R_{fsl}\: [0,\: 4096] \rightarrow Hue_{fsl}\: [45, \: 210]  
\end{cases}
 \label{eq: mapping_reg}
\end{equation}

\begin{equation}
Hue = \frac{Hue_{fsl}\: +\: Hue_{fsr}}{2},
\label{eq: hue}
\end{equation}
where $R_{fsr}$, $R_{fsl}$ are the register values for FSR and FS-L data, respectively; $H_{fsr}$, $H_{fsl}$ are the hue values for FSR and FS-L readings, respectively. 
\newline 

Since the RPi camera measures colour in different value range (i.e. max\:(Hue)\: = \:179, color changes in reversed manner) an inverse transformation is applied:

\begin{equation}
Hue \: [165, \:250] \rightarrow F \: [0,\: 4] ,
\label{eq: mapping_force}
\end{equation}
where $F$ is the estimated force value.

\subsection{Gesture recognition with NN}

To provide a natural remote control during the drone catching task, a custom mocap device V-Arm was developed and integrated into the DroneTrap system. The V-Arm is designed in the form of a glove with 5 flexure sensors to track finger motion. Data processing from sensors is performed with the Arduino Mega controller and then transferred to a control PC for gesture recognition via the UART interface.

\begin{figure}[h!]
  \centering
  \includegraphics[width=0.40\textwidth]{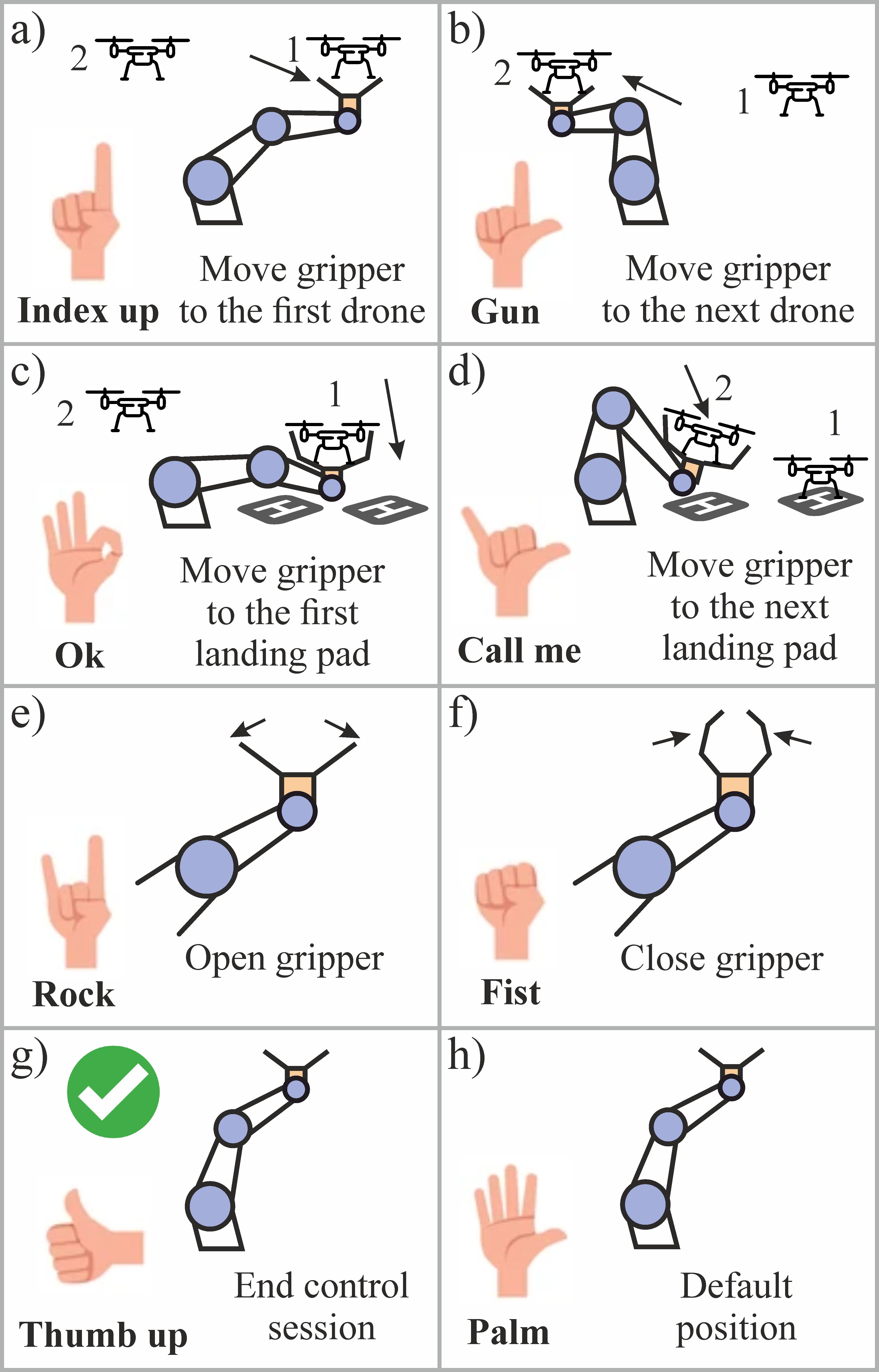}
  \caption{Gestures for NN training: (a) Palm. (b) Gun. (c) Ok. (d) Call me. (e) Rock. (f) Fist. (g) Thumb up. (h) Palm.}
  \label{fig:gestures table}
\end{figure}

The V-Arm must be calibrated for each user before performing gesture recognition. The flexure sensor's digital value is calibrated while the user is performing two gestures: palm open and fist, corresponding to the lowest and highest digital value of the sensor. Thus, the algorithm calibrates the working range according to the anatomy of the user's hand. 
Eight static gestures were chosen for the training process (Fig. \ref{fig:gestures table}).

For the V-arm hand tracking system as raw data, 5 input values corresponding to the thumb, index, middle, ring, and little finger's bending are received in float format and mapped to the 0 to 180-degree range.
As the ground truth device, the LeapMotion Controller was applied to collect the dataset. To obtain robust data by IR sensors of Controller, a dark environment with a monochromatic background was provided. To calculate the bending angle from the LeapMotion data, we propagated the direction vectors of the last and first phalanges of the finger and calculated their intersection point. Raw data of the 5 finger bendings were collected for all 8 gestures from 3 repetitions of each participant.

\begin{figure}[h!]
  \centering
  \includegraphics[width=0.9\linewidth]{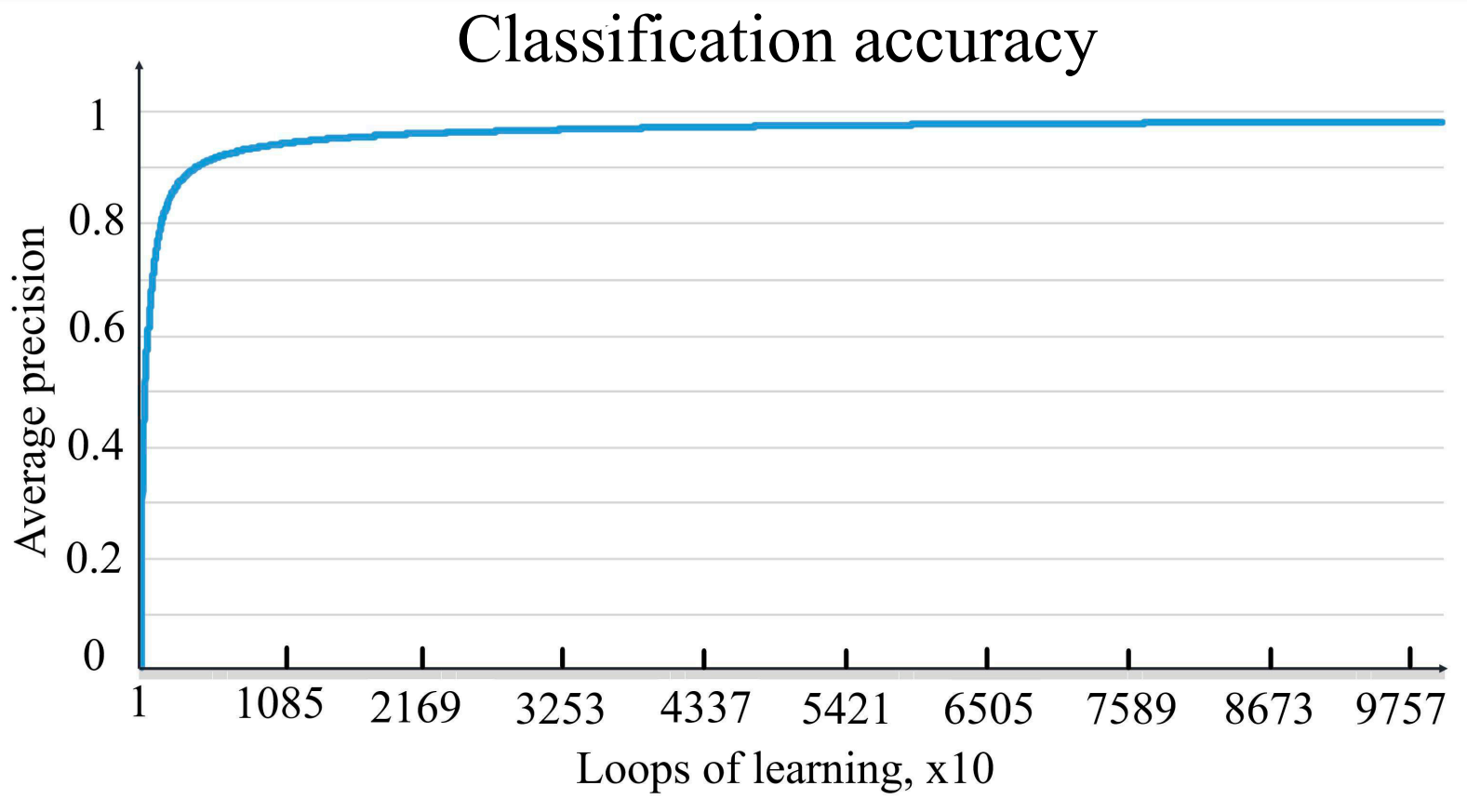}
  \caption{The effect of the number of training loops on the classification accuracy.}
  \label{fig:NN precision graph}
\end{figure}

After the training process, the matching between the set of finger angles and gestures was obtained. 
The dataset consists of 24 gestures. The NN consists of an input vector with 5 values (angles between the palm and each finger), 1 hidden layer with 20 neurons, and an output vector of 8 values (number of gestures), where each set of values corresponds to each gesture. The backpropagation method was applied for the NN training.
The 100 000 training loops with a learning rate of $5.0\cdot10^{-4}$ allowed us to achieve 98.3\% classification accuracy. The average gesture classification accuracy was checked by each 10th loop in training. The output vector was weighted as maximum in case of correct classification and zero if the classification error was revealed. The classification accuracy of trained NN is shown in Fig. \ref{fig:NN precision graph}.

\section{Experimental Evaluation} 
 \subsection{Force estimation}

\subsubsection{Research Methodology}
        
\begin{figure}[h]
  \includegraphics[width=1.0\linewidth]{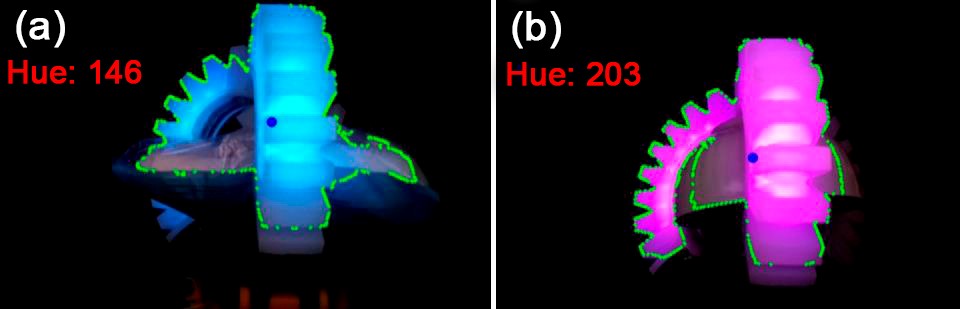}
  \caption{Force detection by the remote color-based system while grasping: a) a soft object, b) a rigid object.}

  \label{fig:2obj}
\end{figure}

The experiment was conducted to evaluate the sensor behavior while grasping two objects of different stiffness: a compliant gauze fabric as the soft object and a solid bowl as the rigid object (Fig. \ref{fig:2obj}). During the experiment, the air supply system applied a minimal pressure of 1.4 kPa for the soft gripper to grasp the object.
The applied force was measured and estimated separately with the CV camera, force, and flex sensor. After reaching 5 s of the full grasping, ASM released pressure from the soft gripper.

\subsubsection{Experimental results}
        
\begin{figure}[h]
  \includegraphics[width=1.0\linewidth]{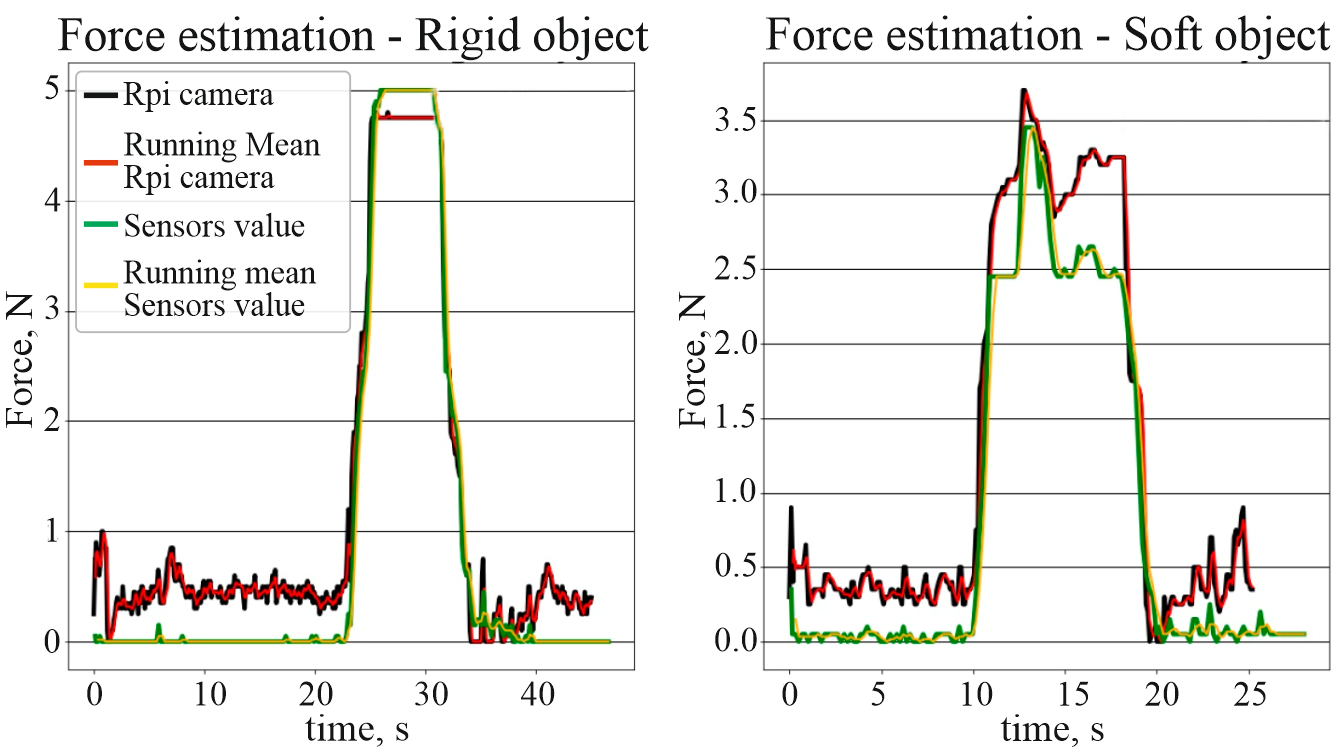}
  \caption{Force estimation by the remote system while grasping: a rigid object (left), a soft object (right).}
  \label{fig:force}
\end{figure}

\begin{figure} [h]
  \includegraphics[width=1.0\linewidth]{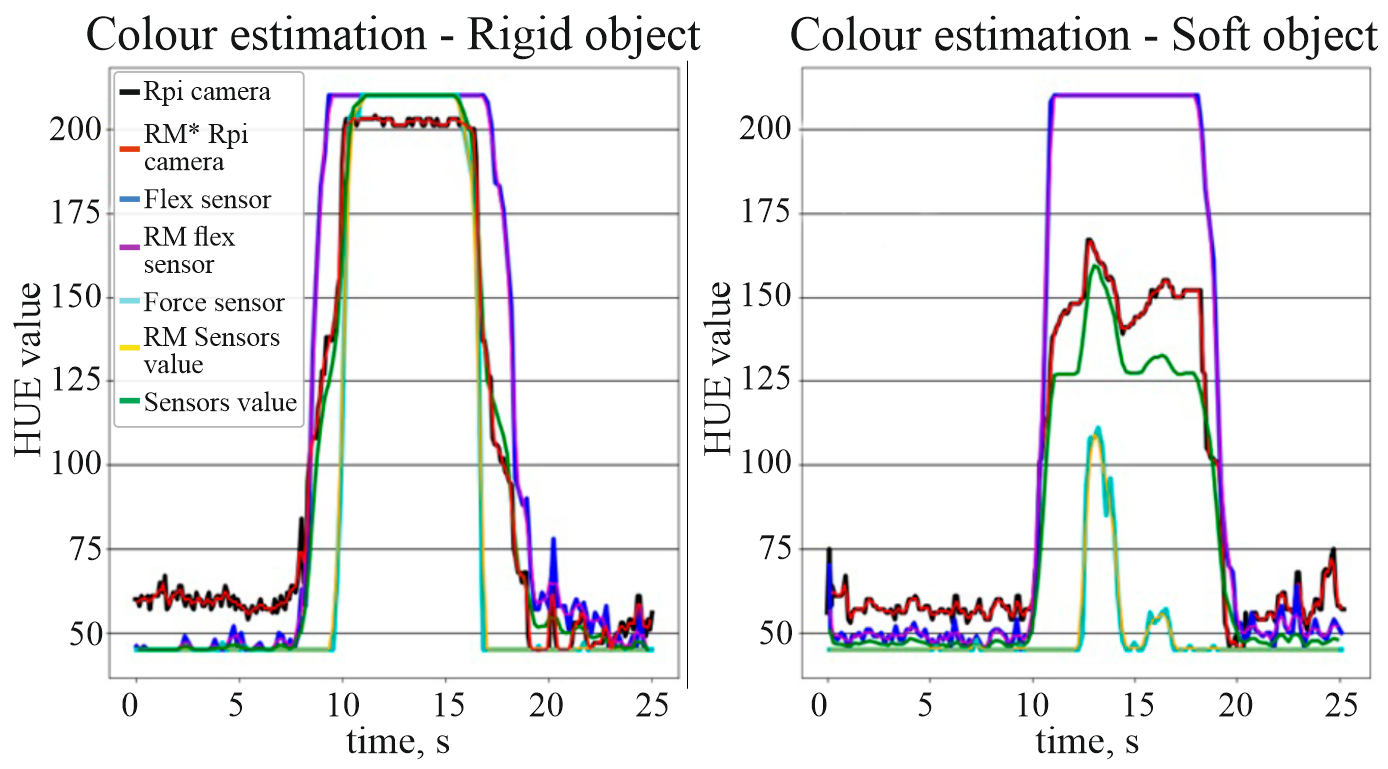}
  \caption{Comparison of soft body with rigid body manipulation in terms of color evaluation: a rigid object (left), a soft object (right).}
  \label{fig:color}
\end{figure}

The experimental results revealed that a soft body grab is performed with 54\% lower mean force than the rigid object grasping (Fig. \ref{fig:force}). 
The estimation error of color and force for the soft object (15.7\%) is much higher than for the rigid one (4.7\%) (Fig. \ref{fig:color}), which was caused by the possibility that a soft object might obscure the view of robotic fingers and prevent the correct light reflection. Therefore, the target's irregular shape impact should be considered when applying the camera color detection algorithm. Considering that a drone is a small object with a low mass and fragile structure, the gripper should operate with drones as a soft body to prevent them from breaking down.

    \subsection{Experiments with gesture recognition using NN}

\subsubsection{System comparison experiment}
To evaluate the developed mocap device (V-Arm), the comparison experiment was conducted with 2 systems: LeapMotion and HTC Vive. In this experiment, we hypothesized that while camera-based approaches may outperform V-Arm in good lighting conditions, the sensor glove's robustness would allow it to achieve better overall results. Therefore, we selected several lighting and hand positioning conditions for the experiment:
\begin{itemize}
\item Well-lightened Room (WL) vs. dim-lightened (DL).
\item Hands showing gestures in front of the headset (0 deg.) vs. turned by 45 deg. vs. turned by 90 deg.
\item Hands placed one above another (HP) vs. hands placed coherently (HNP).
\end{itemize}

For this experiment, 8 participants (mean age = 22) were invited to repeat 8 gestures displayed on the screen with both hands in 5 different environmental conditions. All volunteers previously had experience with motion capture systems. The average recognition rate of the performed gesture samples by the same NN architecture and three different tracking devices is listed in Table \ref{tab:Averaged experiment 1 results}.

\begin{table}[htbp]
\caption{Comparison of 3 Mocap Systems in Various Environments}
  \centering
\begin{tabular}{|l||*{3}{c|}}
\hline
\makebox[30mm]{Conditions} &\makebox[3mm]{Vive }&\makebox[3mm]{Leap}&\makebox[3mm]{V-arm}\\
\hline
WL,   front , HNP & 90.6\% & 84.4\% & 93.8\% \\
\hline
DL,   front , HNP & 68.8\% & 50\% & 90.6\% \\
\hline
WL, 45   deg , HNP & 50\% & 37.5\% & 93.8\% \\
\hline
WL,   profile , HNP & 71.9\% & 75\% & 90.6\% \\
\hline
WL,   front, HP & 68.8\% & 37.5\% & 93.8\%\\
\hline
\end{tabular}
  
  \label{tab:Averaged experiment 1 results}
\end{table}

The experimental results revealed that in a well-lightened room with a frontal hand positioning and absence of hand occlusion, which were considered as the best environmental conditions due to the highest accuracy rate, the V-arm gesture tracking performed on average by 3.2\% better than Vive visual tracking and by 9.4\% better than LeapMotion system. 

However, in the worst lighting condition with 45 deg. hand rotation, V-arm significantly outperformed both camera-based systems (43.8\% in the case of Vive and 56.3\% in the case of LeapMotion) even in the well-lightened room. V-arm showed insensitivity to the environment's lighting conditions as expected, while LeapMotion accuracy decreased by 34.4\% in dim-lightened conditions.

\subsubsection{Gesture recognition experiment}

An additional comparison experiment was performed to evaluate the recognition rate of all the proposed gestures with the V-arm system. In this scenario, the best environment parameters were chosen based on the previous experimental results: well-lightened room, hands placed in front of the headset, and hands located at the same distance from the headset.
For this experiment, 22 participants (mean age = 23) were invited. Each participant was asked to perform 8 gesture samples with both hands 3 times in random order, 24 gestures in total. The combined dataset consists of 1824 gesture data from three devices. The results of the recognition rate in similar environment by 3 systems are listed in Table \ref{tab:results of experiment with gestures}.
 
\begin{figure}[h!]
  \centering
  \includegraphics[width=1.0\linewidth]{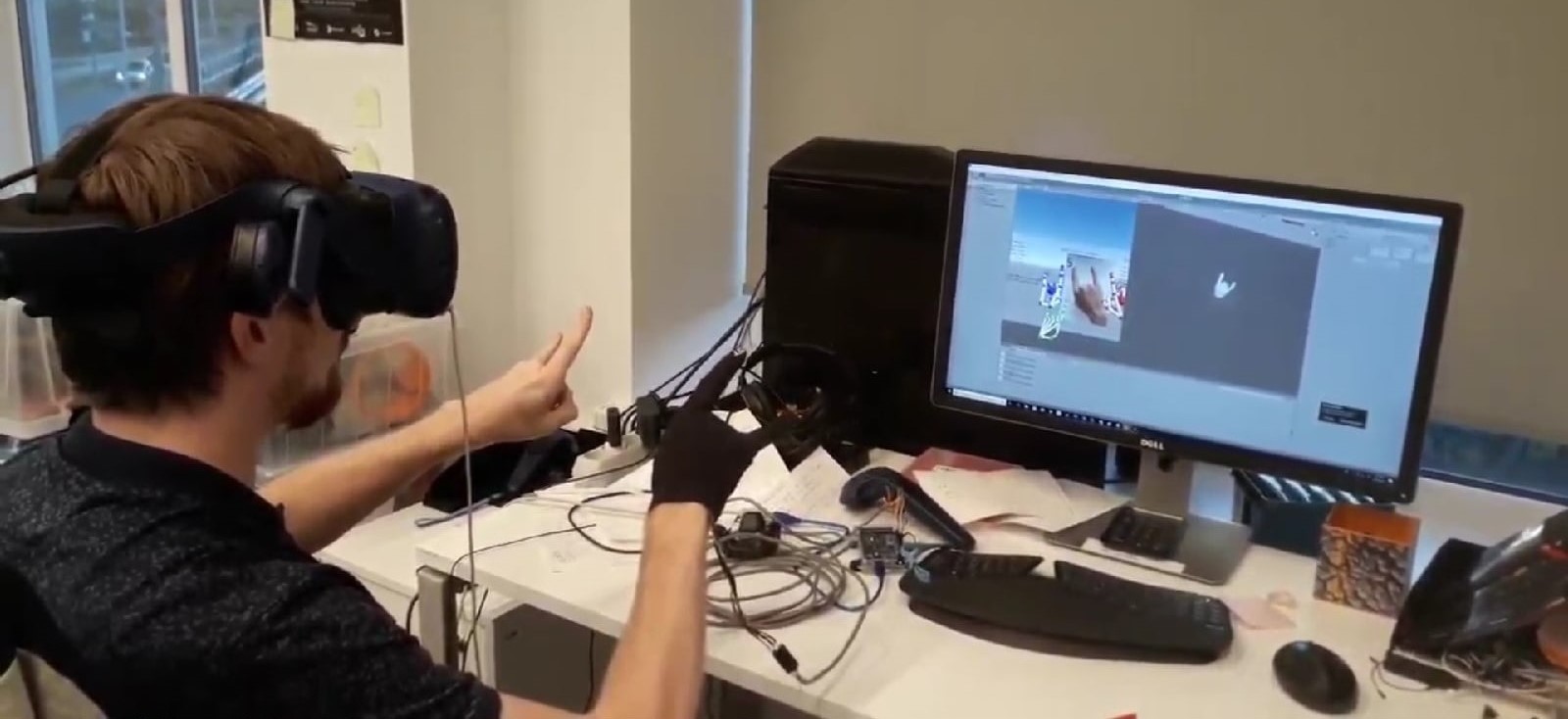}
  \caption{Experimental setup with VIVE headset, LeapMotion camera, and V-arm for the comparison of gesture recognition rate.}
  \label{fig:experiment picture}
\end{figure}





\begin{table}[htbp]
\caption{Comparison Experiment for Gesture Recognition Rate}
\noindent
\centering
\begin{tabular}{|l||*{4}{c|}}\hline
\makebox[20mm]{Gesture}
&\makebox[3em]{Vive}&\makebox[3em]{Leap}&\makebox[3em]{V-Arm}\\\hline

Palm &53.90\%&77.60\%&92.10\%\\\hline
Ok &93.40\%&71.00\%&97.40\%\\\hline
Thumb up &81.60\%&97.40\%&97.40\%\\\hline
Index up &75.90\%&81.60\%&67.10\%\\\hline
Rock &86.80\%&77.60\%&89.50\%\\\hline
Call me &91.10\%&68.40\%&97.40\%\\\hline
Gun &86.80\%&86.80\%&96.00\%\\\hline
Fist &73.70\%&90.80\%&82.90\%\\\hline \hline
Average &78.30\%&81.40\%&90.00\%\\\hline

\end{tabular}

  \label{tab:results of experiment with gestures}
\end{table}%


The experimental results revealed that the average recognition rate of V-Arm was 90\% among all gestures. The V-Arm recognition rate was 11.7\% higher than Vive and 8.6\% higher than LeapMotion systems managed to achieve in the best conditions. According to the obtained data, for the V-Arm device, the most recognizable were “Ok”, “Thumb up” and “Call me” gestures, all with a 97.4\% success rate. 


    

\section{Conclusions}
 
This paper proposes a novel method of safe drone catching based on the developed remote color-based force detection system, hand gesture recognition, and soft robotic gripper with multimodal embedded sensors. Our experimental results revealed that the developed color-based force detection approach can be successfully implemented in catching both rigid (force estimation error 4.7\%) and soft (force estimation error 15.7\%) objects with high precision and provide convenient force estimation either for the autonomous system or for the teleoperation. The operator then can adjust the applied forces and the sequence of swarm catching with the developed gesture recognition system based on the developed mocap glove V-Arm. The experimental results show that the V-Arm achieves a high recognition rate (3.2-56.3\% higher than camera-based systems depending on environment parameters) with various hand gestures (on average 90\% recognition rate).

Therefore, the proposed technology can be potentially implemented for the swarm docking on non-flat surfaces in harsh weather conditions, providing both precise and adjustable grasping process. The remote force detection and gesture recognition allow the operator to intuitively and robustly catch the drones in midair securing safety of multirotor structure. 
In the future, we will explore dynamic swarm-catching scenarios by generating high speed trajectories of the drones.


\addtolength{\textheight}{-1cm}   





\bibliographystyle{IEEEtran}
\bibliography{bib}
\

\end{document}